\documentclass[runningheads]{llncs}

\usepackage{eccv}

\usepackage{eccvabbrv}

\usepackage{graphicx}
\usepackage{booktabs}
\usepackage{makecell}
\usepackage[accsupp]{axessibility}  % Improves PDF readability for those with disabilities.

\usepackage{bm}
\usepackage{hyperref}

\usepackage{orcidlink}

\begin{document}

\title{PhysDrape: Learning Explicit Forces and Collision Constraints for Physically Realistic Garment Draping} 

\titlerunning{PhysDrape}

\author{Minghai Chen\inst{1,2} \and
Mingyuan Liu\inst{1} \and
Ning Ma \inst{1} \and
Jianqing Li \inst{2} \and
Yuxiang Huan\inst{1}}

\authorrunning{Chen et al.}

\institute{Guangdong Institute of Intelligence Science and Technology, China \and
Macau University of Science and Technology, Macau SAR, China \\
\email{\{chenminghai, liumingyuan, huanyuxiang\}@gdiist.cn}}

\maketitle

\begin{figure}
    \centering
    \includegraphics[width=0.95\linewidth]{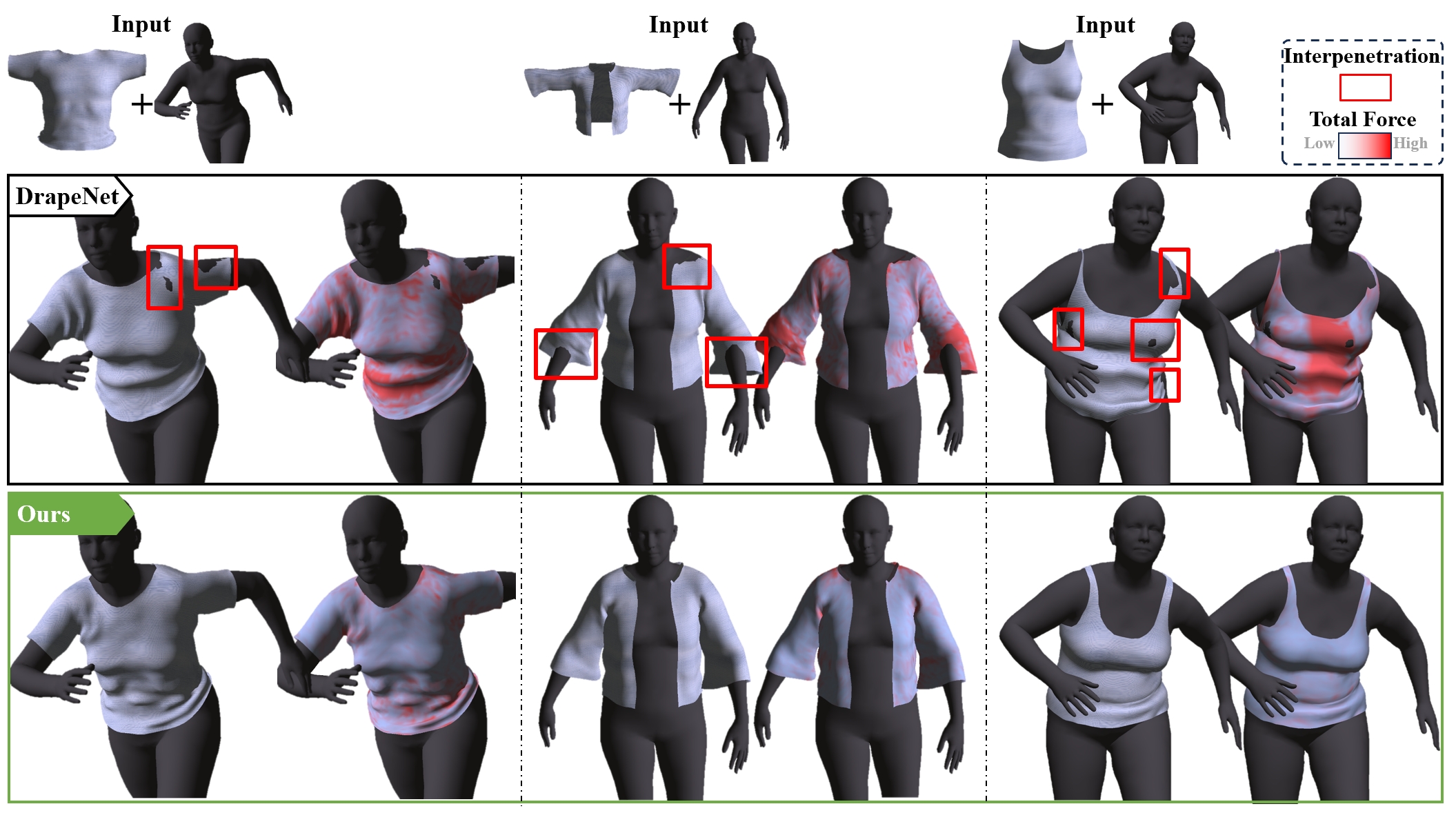}
    \caption{Visualization results show that PhysDrape achieves physically realistic 3D garment draping by unifying deep learning generalization with physical solver fidelity. 
    Across different human poses and garment types, PhysDrape demonstrates quantitatively more plausible force distributions (highlighted in the heatmap) and qualitatively more realistic deformations, with fewer interpenetrations (highlighted in red boxes).}
    \label{fig:placeholder}
\end{figure}

\begin{abstract}
Garment draping aims to fit clothing onto 3D human models. Existing physics-based methods are computationally expensive and struggle to integrate with general differentiable systems. 
In contrast, deep learning-based methods often require extensive annotations and lack explicit physical guarantees, limiting their ability to model accurate details. 
We innovate by bridging physical models with deep learning and propose PhysDrape. 
It integrates three modules using forces as an intermediary: the force-driven GNN predicts forces at each node, the stretching solver models physical deformation, and the collision handler penalizes interpenetration. 
Each module includes learnable parameters to fit underlying force propagation processes and physical properties, enabling generalization to unseen templates and control over properties like stiffness.
PhysDrape enables end-to-end optimization, leveraging intrinsic garment energy for physically plausible deformation via self-supervised learning.
Following the previous training and evaluation protocol on CLOTH3D dataset, PhysDrape achieves lower energy scores, negligible interpenetration, more realistic visualizations, and comparable time consumption.

% Deep learning-based garment draping struggles to simultaneously guarantee strict physical validity and robust collision handling. Existing methods typically rely on soft penalties—creating an intrinsic trade-off between structural distortion and interpenetration—or naive geometric post-processing that sacrifices physical realism. To resolve this, we present PhysDrape, a hybrid neural-physical framework integrating neural inference with explicit geometric solvers. Specifically, a Physics-Informed Graph Neural Network predicts candidate displacements, which are refined by a tightly coupled, differentiable two-stage solver. First, a learnable Force Solver resolves unbalanced Saint Venant-Kirchhoff (StVK) forces to achieve quasi-static equilibrium. Second, a Differentiable Projection layer strictly enforces collision constraints. Unlike disconnected post-processing, embedding this projection within the end-to-end learning loop forces the network to learn collision-aware priors. Extensive experiments demonstrate PhysDrape achieves state-of-the-art performance on unseen garments, ensuring negligible interpenetration $(B2G \approx 0.05\%)$ and significantly lower strain energy than baselines.
\end{abstract}

\section{Introduction}
\label{sec:intro}

Garment draping involves fitting clothing onto a 3D human model in various poses to achieve realistic body conformity. 
This process simulates fabric deformations such as folding, stretching, and motion, which is essential for applications like virtual try-on, animation, gaming, and the metaverse \cite{islam2024deep, xi2021shopping, shi2025generative}.

Existing approaches primarily rely on physics-based simulations or deep learning-based regression, while each has its limitations.
Specifically, \textbf{physics-based simulations} employ multi-round solvers to model fabric properties, incurring high computational costs \cite{baraff2023large, liu2017quasi, provot1997collision, provot1995deformation, su2018gpu, vassilev2001fast, zeller2005cloth, MarvelousDesigner, nvidia2018nvcloth, nvidia2018flex, optitex2018}. 
Additionally, their reliance on discrete computation hinders joint optimization with other algorithms, such as recovering 3D shapes and physical parameters from visual input, limiting their generalization across diverse scenarios.
\textbf{Deep learning-based regression} methods learn from large annotated datasets and directly predict garment geometry from body poses \cite{gundogdu2019garnet, gundogdu2020garnet++, patel2020tailornet, bertiche2021deepsd, santesteban2019learning, bhatnagar2019multi}. 
However, these methods lack explicit physical constraints, leading to unrealistic garment behavior, e.g., unnatural draping, interpenetration, and inaccurate garment details.

Although pioneering efforts integrate deep learning with physics-informed designs, they fall short of explicitly incorporating physical processes into deep learning models, and thus lack a guarantee of accurate physical interaction between the garment and the body.
In particular, earlier works \cite{bertiche2020pbns, santesteban2022snug} primarily focus on incorporating physics-informed loss functions specifically tailored to individual garments.
Subsequent works \cite{de2023drapenet, li2023isp, grigorev2023hood, liao2024senc, tiwari2023gensim} aim to generalize the model across diverse garment topologies.
Additionally, supervised postprocessors \cite{tan2022repulsive, tang2025saf} can be applied after coarse draping to rectify physically inconsistent regions.
However, these methods face two main issues: 1) modeling physical constraints as loss functions lacks the necessary physical guarantees, leading to inaccurate results such as severe interpenetration. 2) postprocessing fails to jointly optimize with garment draping, leading to suboptimal fabric deformation.

% In response to these challenges, physics-based self-supervision has emerged as a promising direction, constraining predictions to obey physical laws without requiring ground-truth data. Pioneering works \cite{bertiche2020pbns, santesteban2022snug} introduced optimization frameworks tailored for specific garments, while subsequent methods \cite{de2023drapenet, li2023isp, grigorev2023hood, liao2024senc, tiwari2023gensim} extended this capability to generalize across diverse garment topologies. Despite their success, a fundamental limitation persists across these neural solvers: they predominantly enforce physical validity through soft penalty terms in the loss function. This formulation induces an intrinsic trade-off: aggressively penalizing collisions often leads to geometric distortion and non-physical stretching, whereas prioritizing structural preservation inevitably permits interpenetration. Consequently, soft-constrained formulations struggle to guarantee both strict collision constraint and physical fidelity simultaneously. While learned correctors \cite{tan2022repulsive, tang2025saf} attempt to address this via predictive offsets, they rely on ground-truth supervision and lack explicit solvers to ensure physical equilibrium.

In this work, we propose PhysDrape, which explicitly integrates deep learning models with physical deformation models. 
It consists of three components: force-driven GNN, stretching solver, and collision handler.
Specifically, the force-driven GNN estimates the forces acting on each node of the mesh, rather than the typical position regression, enabling enhanced mesh-based physical deformation that adapts to complex body geometries.
The stretching solver iteratively deforms the garment by physically propagating forces through the mesh, with learnable parameters fitting the force propagation patterns in the fabric.
The collision handler pushes the fabric out of the body to resolve interpenetration, while using learnable parameters to fit hard-to-model physical quantities, such as elasticity, at the body-garment interface.
Notably, these three modules are learnable and enable end-to-end optimization in a self-supervised manner, using an energy-based loss function. 
This joint optimization of the neural network and explicit learnable physical solver not only enables generalization to unseen garments, but also allows control over physical parameters, such as stiffness, facilitating more physically plausible draping.
Following previous protocols \cite{de2023drapenet}, PhysDrape is validated using garments and template meshes from the CLOTH3D dataset, without ground-truth \cite{bertiche2020cloth3d}.
The results demonstrate that PhysDrape achieves better energy scores, fewer interpenetration, more realistic visualizations, and comparable time consumption.

% In this work, we present PhysDrape, a hybrid neural-physical solver designed to bridge the gap between neural efficiency and explicit physical constraints. Unlike frameworks that implicitly trade off geometric feasibility against physical plausibility, PhysDrape tightly couples a Graph Neural Network (GNN) with a learnable explicit solver. By integrating a learnable Force Solver with a fully differentiable projection layer, we unify quasi-static physical equilibrium and geometric correction into an end-to-end learning framework. Here, explicit internal forces and collision constraints serve as differentiable supervision to enforce physically plausible results.

To summarize, our main contributions are threefold:  
\begin{itemize} 
\item We propose PhysDrape, a novel model that unifies neural networks and physical simulations through forces as an intermediary, enabling self-supervised end-to-end optimization for more physically realistic 3D garment draping.
% \item We propose PhysDrape, a hybrid framework that resolves the conflict between collision handling and structural preservation by integrating neural inference with explicit physics and geometric solvers in a differentiable loop. 
\item We propose a novel differentiable stretching and collision solver, integrated with learnable parameters to fit hard-to-measure physical quantities for physical fidelity. It enables joint optimization with neural networks, providing not only generalization to unseen garments but also explicit control over physical stiffness to simulate different fabrics.
% \item We design a Physics-Informed GNN to predict candidate deformations, which are refined by a two-stage solver: a learnable Force Solver that minimizes Saint Venant-Kirchhoff (StVK) residual forces, and a differentiable projection layer that strictly enforces collision constraints. 
\item PhysDrape achieves state-of-the-art performance, reflected in lower energy scores, less interpenetration, more realistic visualizations, and comparable time consumption. Ablation experiments further demonstrate the effectiveness of each design. 
\end{itemize}

\section{Related Works}
\label{sec:relate}
\subsection{3D Garment Draping}

Approaches for garment draping generally fall into two categories: physics-based simulation and deep learning-based regression. 
Physics-based simulation models cloth dynamics using Mass–Spring Systems (MSS) \cite{provot1995deformation} or Finite Element Methods (FEM) \cite{etzmuss2003deriving}, typically employing optimization-based solvers like Position-Based Dynamics (PBD) \cite{muller2007position}, Extended PBD (XPBD) \cite{macklin2016xpbd}, and Projective Dynamics (PD) \cite{bouaziz2023projective} for stability. 
\textit{However}, these methods typically rely on non-differentiable discrete projections, limiting joint optimization with other approaches and thus hindering their application in modern AI systems. While differentiable variants like DiffXPBD \cite{stuyck2023diffxpbd} and IPC \cite{li2020incremental} exist, they require large linear systems at each step, imposing heavy computational burdens. As a result, differentiable deep learning systems are increasingly emphasized.

Deep learning-based 3D garment draping focuses on regressing garment geometry from body poses, later incorporating physics-informed designs to improve realism.
Early works generate large datasets of body-garment pairs through annotations and offline simulations to train neural networks in a supervised manner \cite{gundogdu2019garnet, gundogdu2020garnet++, patel2020tailornet, bertiche2021deepsd, santesteban2019learning, bhatnagar2019multi}. These approaches are labor-intensive and suffer from a significant sim-to-real gap.
Latter works introduce energy-based loss functions to train neural networks, guiding the garment draping results to adhere to physical laws from an optimization perspective.
For instance, SNUG \cite{santesteban2022snug} and PBNS \cite{bertiche2020pbns} pioneer physics-based self-supervision by directly minimizing physical energies. 
Santesteban et al. \cite{santesteban2021self} propose a generative framework that utilizes a diffused body representation to explicitly handle collisions.
GAPS \cite{chen2024gaps} introduces a collision-aware geometrical loss, allowing adaptive stretching to resolve interpenetrations.
To generalize to more complex and diverse garment templates, later works optimize for various garment types and layers. DIG \cite{li2022dig} introduces implicit Signed Distance Functions (SDF) \cite{park2019deepsdf} for draping, handling arbitrary geometries and reconstructing meshes with the marching cubes algorithm \cite{lewiner2003efficient}. 
DrapeNet \cite{de2023drapenet} integrates physics-based objectives into a generative framework based on unsigned distance functions \cite{chibane2020neural, venkatesh2021deep, zhao2021learning}, enabling realistic draping of diverse fabrics.
ISP \cite{li2023isp} further introduces implicit sewing patterns for complex multi-layered garments. 
\textit{However}, these methods are limited by soft penalties at the loss function level and often rely on postprocessing to handle collisions instead of joint optimization, leading to a lack of physical guarantees and suboptimal model optimization. In contrast, our PhysDrape constructs explicit physical processes at the model level and seamlessly integrates the neural network, stretching solver, and collision handler into an end-to-end optimizable framework, further ensuring the realism of 3D garment draping. 
Additionally, the physical-neural integration allows the model to generalize to unseen garments and support the simulation of garment with manually specified stiffness.

\subsection{Garment-Body Interpenetration}

To prevent the garment from penetrating the body geometry, existing methods primarily rely on geometric correction and penalty-based learning approaches.
Geometric correction approaches \cite{santesteban2019learning} directly project penetrating vertices onto the body surface. This process is non-differentiable, ignores physical constraints, and hence leads to non-physical distortions.
Penalty-based learning is commonly used in neural network-based methods and enables end-to-end training. For example, \cite{bertiche2020pbns, santesteban2022snug, de2023drapenet} employ soft collision penalties, with SENC \cite{liao2024senc} introducing a global intersection volume loss for self-collisions. 
ReFU \cite{tan2022repulsive} adds a learnable unit to predict displacement scales for SDF-based repulsion and SAF \cite{tang2025saf} extends this correction to the triangle mesh using local features.
\textit{However}, the aforementioned methods rely on collision postprocessors, novel loss optimizations, and annotated labels for collision avoidance learning, which often result in laborious annotations and unresolved garment force imbalances. In contrast, PhysDrape self-supervisedly learns and explicitly integrates a differentiable design into the learning system, avoiding extensive annotations while jointly optimizing penetration and fabric deformation to ensure physical fidelity.

\section{Methods}
\label{sec:methods}
\subsection{Overview}

\begin{figure}[h]
    \centering
    \includegraphics[width=0.95\linewidth]{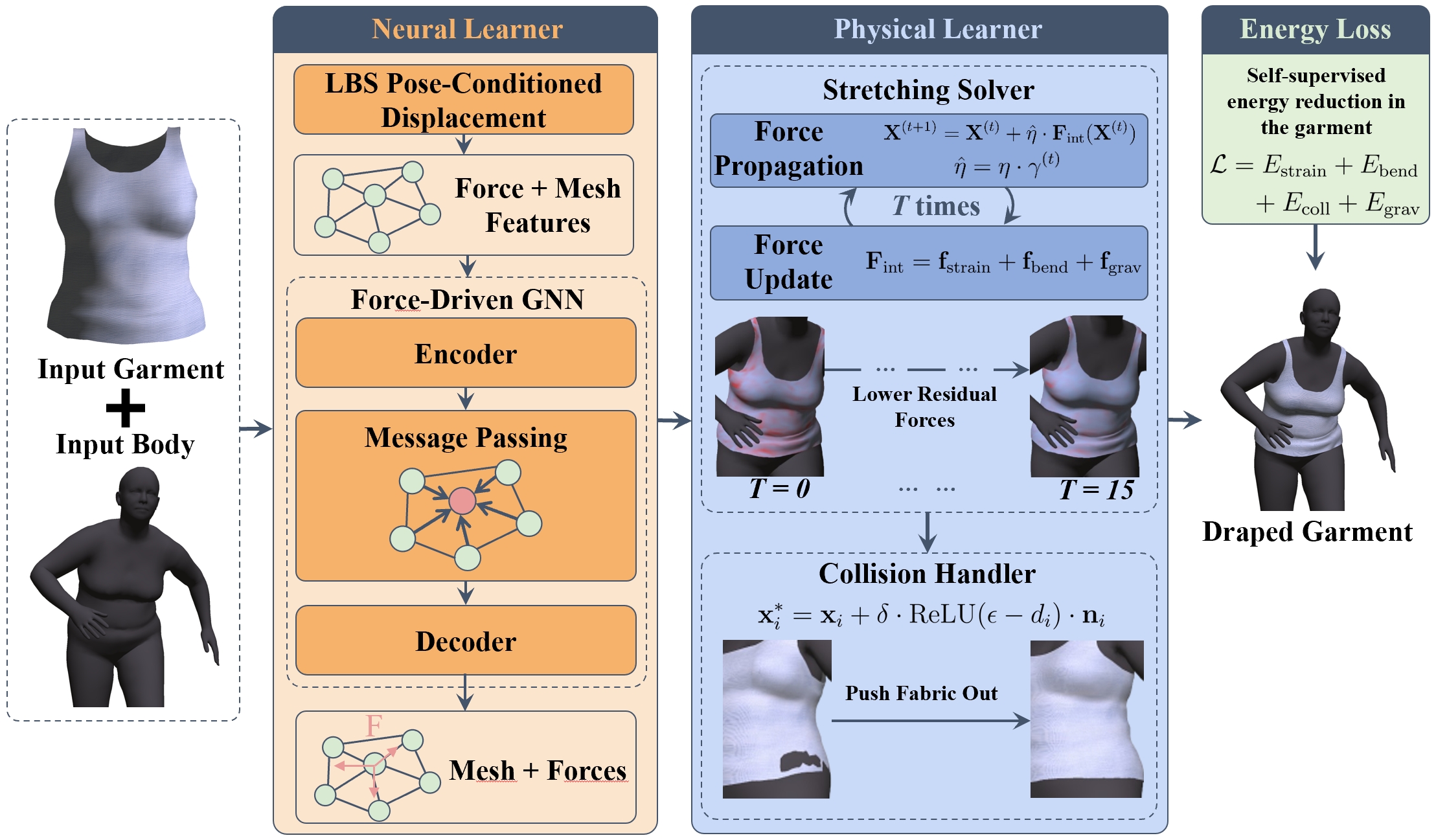}
    \caption{PhysDrape innovatively unifies neural networks and explicit physical simulations through forces as an intermediary, enabling self-supervised end-to-end optimization for more physically realistic 3D garment draping. It integrates a force-driven GNN, a stretching solver, and a collision handler, with learnable parameters to model unmeasurable physical properties, such as intrinsic fabric properties and body-garment interaction, ensuring superior physical fidelity. }
    \label{fig:overview}
\end{figure}

We propose PhysDrape, which integrates explicit physical processes into deep learning models to achieve physically realistic 3D draping, as illustrated in Fig. \ref{fig:overview}. 
PhysDrape, centered around forces acting on garment mesh nodes (Sec. \ref{sec3-2}), consists of three main components: the force-driven GNN, the stretching solver, and the collision handler.
\textbf{The force-driven GNN} (Sec. \ref{sec3-3}) produce a garment and derives the physical forces on the garment mesh by considering both human pose and garment shape, ensuring superior physical fidelity.
\textbf{The stretching solver} (Sec. \ref{sec3-4}) deforms the garment based on the derived forces from GNN, with learnable factors fitting the properties of force propagation through the fabric.
\textbf{The collision handler} (Sec. \ref{sec3-5}) resolves overlaps and introduces learnable contact parameters, representing the physical quantities of body-garment contact.
All three modules are differentiable, enabling end-to-end optimization of the entire system using self-supervision (Sec. \ref{sec3-6}), resulting in more realistic deformation. We adopt the SMPL \cite{loper2023smpl} model to parameterize human body mesh. To formalize the internal fabric mechanics, we adopt the Saint Venant-Kirchhoff (StVK) material model \cite{narain2012adaptive}, which is widely proven effective in virtual garment simulation and neural garment modeling \cite{santesteban2022snug, de2023drapenet, grigorev2023hood}.

\subsection{Draping Force Formulation} \label{sec3-2}

To achieve physically plausible draping, we formulate the draping process as the pursuit of a quasi-static state by explicitly resolving intrinsic force in garment. Specifically, the draped state is governed by three distinct components: (1) the strain force $\mathbf{f}_{\text{strain}}$, which penalizes in-plane stretching and shearing to preserve the static structure of garment (2) the bending force $\mathbf{f}_{\text{bend}}$, which resists out-of-plane sharp creases to maintain surface smoothness; and (3) the gravitational force $\mathbf{f}_{\text{grav}}$, which naturally drives the drape downwards. Under the StVK material formulation, we model the in-plane strain force acting on vertex $i$ as:

\begin{equation}
\label{eq:forces_strain}
\begin{aligned}
    &\mathbf{f}_{\text{strain}, i} = -\sum\nolimits_{T \in \mathcal{N}(i)} V_T \mathbf{P}_T \mathbf{b}_{i,T} \\
    %&\mathbf{f}_{\text{grav}}   = m_i \mathbf{g},
\end{aligned}
\end{equation}
, where $V_T$ denotes the reference volume (i.e., rest area multiplied by thickness) of the mesh face $T$. $\mathbf{P}_T$ is the First Piola-Kirchhoff stress tensor, derived as $\mathbf{P}_T = \mathbf{F}_T \mathbf{S}_T$, where $\mathbf{F}_T$ is the deformation gradient mapping the rest pose to the current state, and $\mathbf{S}_T = 2\mu \mathbf{G} + \lambda \text{Tr}(\mathbf{G})\mathbf{I}$ is the second Piola-Kirchhoff stress based on the Green-Lagrange strain $\mathbf{G}$ \cite{narain2012adaptive}. The Lamé constants $\mu$ and $\lambda$ govern the fabric's shear and stretch stiffness, respectively. Finally, $\mathbf{b}_{i,T}$ represents the shape function gradient evaluated at the rest pose, serving as a geometric mapping that distributes the face-wise stress back to the specific vertex $i$ as a directional nodal force.

To penalize sharp out-of-plane creases, the bending force on vertex $i$ is defined as: 
\begin{equation}
\label{eq:forces_bend}
\begin{aligned}
    & \mathbf{f}_{\text{bend}, i} = - \sum\nolimits_{e \in \mathcal{H}(i)} k_{\text{bend}} s_e \theta_e \nabla_{\mathbf{x}_i} \theta_e \
\end{aligned}
\end{equation}
, where $\mathcal{H}(i)$ denotes the set of all local hinges $e$ (formed by adjacent triangle pairs) that involve vertex $i$; $k_{\text{bend}}$ is the bending stiffness coefficient; and $s_e = l_e^2 / (4A_e)$ acts as the geometric scale factor of the hinge. The term $\nabla_{\mathbf{x}_i} \theta_e$ represents the gradient of the dihedral angle $\theta_e$ with respect to the vertex $i$, inherently directing the restoring force along the face normals to penalize bending deformation. 

Finally, to naturally drive the fabric downwards, the gravitational force is given by: 
\begin{equation}
\label{eq:forces_grav}
\begin{aligned}
    &\mathbf{f}_{\text{grav}, i}   = m_i \mathbf{g}
\end{aligned}
\end{equation}
, where $m_i$ denotes the mass assigned to vertex $i$, and $\mathbf{g}$ represents the gravitational acceleration vector.

The internal active force $\mathbf{f}_{\text{int}, i}$ acting on vertex $i$ is the aggregation of all driving forces:

\begin{equation}
\label{eq:force_int}
\mathbf{f}_{\text{int}, i} = \mathbf{f}_{\text{strain}, i} + \mathbf{f}_{\text{bend}, i} + \mathbf{f}_{\text{grav}, i}
\end{equation}

PhysDrape explicitly learns to propagate and resolve these forces acting on the garment to achieve force equilibrium and produce physically plausible draping. 

\subsection{Force-Driven Graph Neural Network} \label{sec3-3}

To model the physical interaction between the human body and the fabric, we introduce the Force-Driven Graph Neural Network (GNN) to derive the the garment mesh and the physical forces acting upon the garment. To initialize this process, we follow established practices \cite{santesteban2021self, li2022dig} to generate a pose- and shape-conditioned deformation over the template garment $\mathbf{X}_{\text{temp}} \in \mathbb{R}^{N \times 3}$, roughly fitting it to the target human body:

\begin{equation}
\label{eq:coarse_init}
    \mathbf{X}_{\text{init}}  = \mathcal{W}(\mathbf{X}_{\text{shape}}, \bm{\beta}, \bm{\theta}, \mathcal{W}_G), where~
    \mathbf{X}_{\text{shape}} = \mathbf{X}_{\text{temp}} + \Delta \mathbf{X}_{\text{shape}}(\beta)
\end{equation}
, where $\Delta \mathbf{X}_{\text{shape}}(\beta) \in \mathbb{R}^{N \times 3}$ represents the shape-dependent vertex displacements driven by the human body shape parameters $\bm{\beta} \in \mathbb{R}^{10}$. The function $\mathcal{W}(\cdot)$ denotes the standard Linear Blend Skinning (LBS) algorithm \cite{loper2023smpl} parameterized by the target pose $\bm{\theta} \in \mathbb{R}^{72}$. Following \cite{santesteban2021self, li2022dig}, both the shape displacements $\Delta \mathbf{X}_{\text{shape}}$ and the skinning weights $\mathcal{W}_G$ are formulated as continuous learned functions of the target body parameters to enable differentiable kinematic fitting. While this kinematic initialization $\mathbf{X}_{\text{init}}$ efficiently aligns the garment with the body pose, it completely lacks physical awareness, often resulting in unnatural artifacts and body-garment interpenetration. Therefore, we formulate the garment mesh as a graph $\mathcal{G} = (\mathcal{V}, \mathcal{E})$ based on its underlying topology. Using $\mathbf{X}_{\text{init}}$ as the geometric foundation, we explicitly compute the node features $\mathbf{h}_i$ for each vertex $i \in \mathcal{V}$ and the edge features $\mathbf{e}_{ij}$ for each connected pair $(i, j) \in \mathcal{E}$ as follows:

\begin{equation}
\label{eq:features}
\begin{aligned}
    \mathbf{h}_i &= [\mathbf{n}_i, \mathbf{d}_i, \mathbf{f}_{\text{int}, i}, \mathbf{m}_i] \in \mathbb{R}^{11}, \\
    \mathbf{e}_{ij} &= [\mathbf{x}_{ij}, \|\mathbf{x}_{ij}\|, \mathbf{x}_{ij}^0, l_{ij}^0] \in \mathbb{R}^{8}
\end{aligned}
\end{equation}

\subsubsection{Node Features.} The node feature $\mathbf{h}_i$ equips the GNN with both geometric context and physical states. Specifically, $\mathbf{n}_i$ is the vertex normal for capturing local surface orientation and curvature, and $\mathbf{d}_i$ denotes the signed distance to the human body surface, enabling the network to perceive potential collisions. Crucially, we explicitly compute the internal active force $\mathbf{f}_{\text{int}, i}$ (as defined in Eq. \ref{eq:force_int}) at the current initialized state $\mathbf{X}_{\text{init}}$. Feeding this force into the GNN provides the network with a direct physical gradient toward equilibrium. Finally, $\mathbf{m}_i = [\mu, \lambda, k_{\text{bend}}, m_i]$ encodes the vertex-specific material properties, comprising the Lamé coefficients for membrane stiffness, the bending coefficient, and the nodal mass.

\subsubsection{Edge Features.} To capture local topological deformations, the edge feature $\mathbf{e}_{ij}$ concatenates geometric properties in both the current deformed space ($\mathbf{x}_{ij} = \mathbf{x}_i - \mathbf{x}_j$ and its norm $\|\mathbf{x}_{ij}\|$) and the canonical rest space ($\mathbf{x}_{ij}^0$ and the rest length $l_{ij}^0$). This dual-space representation explicitly allows the message passing layers to measure local stretch and orientation variations relative to the rest state.

\subsubsection{Architecture.} GNN in PhysDrape is an encode-process-decode architecture \cite{pfaff2020learning}. First, the Encoder module projects the raw node features $\mathbf{h}_i$ and edge features $\mathbf{e}_{ij}$ (defined in Eq.~\ref{eq:features}) into high-dimensional latent embeddings $\mathbf{h}_i^{(0)}$ and $\mathbf{e}_{ij}^{(0)}$, respectively. This is achieved via two independent multi-layer perceptrons, denoted as $\text{MLP}_{\text{node}}^{\text{enc}}(\cdot)$ and $\text{MLP}_{\text{edge}}^{\text{enc}}(\cdot)$, respectively: 

\begin{equation}
\begin{aligned}
    \mathbf{h}_i^{(0)} &= \text{MLP}^{\text{enc}}_{\text{node}}(\mathbf{h}_i), \\
    \mathbf{e}_{ij}^{(0)} &= \text{MLP}^{\text{enc}}_{\text{edge}}(\mathbf{e}_{ij})
\end{aligned}
\end{equation}

Then, the processor iteratively updates these latent representations via $L$ message passing layers to propagate physical interactions across the garment topology. Specifically, at each layer $l$, an edge network $\text{MLP}_{\text{edge}}(\cdot)$ updates the interaction features between connected vertices, and a node network $\text{MLP}_{\text{node}}(\cdot)$ subsequently updates the individual vertex states by aggregating messages from their local neighborhoods $\mathcal{N}(i)$:

\begin{equation}
\label{eq:mpnn_update}
\begin{aligned}
    \mathbf{e}_{ij}^{(l+1)} &= \mathbf{e}_{ij}^{(l)} + \text{MLP}_{\text{edge}}\left( \mathbf{e}_{ij}^{(l)}, \mathbf{h}_i^{(l)}, \mathbf{h}_j^{(l)} \right), \\
    \mathbf{h}_i^{(l+1)} &= \mathbf{h}_i^{(l)} + \text{MLP}_{\text{node}}\left( \mathbf{h}_i^{(l)}, \sum\nolimits_{j \in \mathcal{N}(i)} \mathbf{e}_{ij}^{(l+1)} \right)
\end{aligned}
\end{equation}

Finally, a decoder $\text{MLP}^{\text{dec}}(\cdot)$ projects the evolved node embeddings $\mathbf{h}_i^{(L)}$ back into the physical space to produce the final network predictions:
\begin{equation}
    \label{eq:gnn_pred}
    \hat{\mathbf{X}}_{\text{pred}, i}, \hat{\mathbf{f}}_{i} = \text{MLP}^{\text{dec}}(\mathbf{h}_i^{(L)})
\end{equation}

Notably, rather than merely predicting geometric displacements, our GNN simultaneously infers the garment mesh and the forces acting upon it.

\subsection{Learnable Stretching Solver} \label{sec3-4}
We employ a learnable stretching solver $\mathcal{S}_{\text{force}}$ to propagate the forces acting upon the garment mesh and to relax the garment into an equilibrium state. 
Starting from the output of the force-driven GNN, the solver performs $T$ explicit forward simulation steps to propagate forces and update the garment mesh.
Specifically, at each iteration $t \in [0, T-1]$, we calculate the updated internal active forces by evaluating $\mathbf{F}_{\text{int}}(\cdot)$ at the current mesh geometry $\mathbf{X}^{(t)}$, following the mechanical formulations defined in Sec. \ref{sec3-2}. Subsequently, the vertex positions are updated along the direction of these forces to propagate the physical stretch:

\begin{equation}
\label{eq:solver_update}
    \mathbf{X}^{(t+1)} = \mathbf{X}^{(t)} + \hat{\eta} \cdot \mathbf{F}_{\text{int}}(\mathbf{X}^{(t)}), \quad \hat{\eta} = \eta \cdot \gamma^{(t)}
\end{equation}
, where $\eta$ is the learnable compliance parameter (inverse stiffness), and $\gamma \in (0,1)$ is a decay rate to stabilize the force solving process as the system approaches equilibrium.

\subsection{Collision Handler} \label{sec3-5}

To enforce the non-penetration constraint, we pass $\mathbf{X}^{(T)}$ from the stretching resolver to a differentiable collision handler $\mathcal{P}_{\text{proj}}$. 
For each candidate vertex $\mathbf{x}_i \in \mathbf{X}^{(T)}$, we identify its nearest neighbor $\mathbf{p}_i$ on the body surface and the corresponding outward surface normal $\mathbf{n}_i$. The signed distance from the garment vertex to the body is computed as $d_i = (\mathbf{x}_i - \mathbf{p}_i) \cdot \mathbf{n}_i$.

A collision occurs if this distance falls below a predefined threshold ($d_i < \epsilon$). We actively resolve such penetrations by projecting the vertex outward along the body normal $\mathbf{n}_i$:
\begin{equation}
\label{eq:projection}
\mathbf{x}^*_i = \mathbf{x}_i + \delta \cdot \text{ReLU}(\epsilon - d_i) \cdot \mathbf{n}_i
\end{equation}
, where $\mathbf{x}^*_i$ represents the final collision-free vertex position, and $\epsilon$ is the margin. 
To guarantee non-penetration, we introduce a learnable projection scalar $\delta \ge 1$. By adaptively pushing vertices further outward, $\delta$ creates a geometric buffer that ensures the attached continuous fabric faces are lifted completely clear of the body.
Unlike taking a collision avoidance as post-processing \cite{santesteban2019learning}, our projection is embedded within the end-to-end learning pipeline. This integration ensures that the force-driven GNN and the stretching solver receive direct gradient feedback from the collision constraints, driving the entire network to minimize internal forces and seek physical equilibrium strictly within a valid, interpenetration-free state space.

\subsection{Training Objective} \label{sec3-6}

PhysDrape is trained in a self-supervised manner, eliminating the need for ground-truth 3D draped garments, which enhances its generalization capability. We use the energy-based loss function proposed in prior work \cite{de2023drapenet} as the optimization objective. Specifically, the overall formulation of the physics-informed training objective is as follows:

\begin{equation}
\mathcal{L} = E_{\text{strain}} + E_{\text{bend}} + E_{\text{coll}} + E_{\text{grav}}
\end{equation}

The total loss consists of four energy terms defined on the garment triangle mesh: in-plane fabric stretching ($E_{\text{strain}}$), out-of-plane sharp foldings ($E_{\text{bend}}$), body-garment collisions ($E_{\text{coll}}$), and gravity ($E_{\text{grav}}$). Together, these terms synthesize a  draping state without requiring annotated 3D supervision.
Building upon this loss function, this work focuses on integrating physical solvers with neural networks to model explicit and learnable physical processes beyond conventional loss optimization.

\section{Experiments and Results}
\label{sec:experiments}

% We first introduce the datasets, evaluation metrics and implementation details. We then present a comparison of quantitative and qualitative draping results with state-of-the-art methods. Subsequently, we provide an in-depth analysis of the proposed learnable solvers, focusing on their convergence behavior and physical compliance. Finally, we verify the necessity of each component through ablation studies.

\subsection{Datasets and Implementation Details}

\subsubsection{Datasets.} The CLOTH3D dataset \cite{bertiche2020cloth3d} is used for training and evaluation, following the experimental protocol in DrapeNet \cite{de2023drapenet}, with a focus on upper-body garments, including t-shirts, shirts, tank tops, etc.
For training, we randomly select 600 top garments and use template meshes in their canonical state (T-pose on an average body shape).
Notably, this work follows a self-supervised design, requiring no additional annotations for training. Therefore, we do not use the simulated deformations of the selected garments from the dataset as supervisory signals.
For testing, we use a separate set of 30 unseen top garments to better assess the generalization ability of the model to different garments.
Body poses are randomly sampled from the AMASS dataset \cite{mahmood2019amass}, and the shape parameter
$\bm{\beta}$ is uniformly sampled from $[-3, 3]^{10}$, consistent with \cite{santesteban2022snug,de2023drapenet}.

%\subsubsection{Datasets.} We follow the experimental protocol established in DrapeNet \cite{de2023drapenet} to utilize the CLOTH3D dataset \cite{bertiche2020cloth3d} for both training and evaluation. However, we focus exclusively on upper-body garments. We randomly select 600 top garments (including t-shirts, shirts, and tank tops) from the dataset. Consistent with~\cite{de2023drapenet}, we do not rely on the ground-truth simulated deformations provided by CLOTH3D for supervision; instead, we only utilize the garment template meshes in their canonical state (T-pose on an average body shape). For evaluation, we employ a separate set of 30 top garments that are unseen during training. To ensure robustness across diverse body configurations, we sample body poses from the AMASS dataset \cite{mahmood2019amass} and sample body shape parameters $\beta$ uniformly from the range $[-3, 3]^{10}$ during training.

\subsubsection{Implementation Details.} 
In PhysDrape, the physics-informed GNN follows an encode-process-decode architecture \cite{pfaff2020learning}. The encoder maps the input features to a 128-dimensional latent space. The processor comprises 16 message-passing blocks, with each block employing a 2-layer MLP with ReLU activation and Layer Normalization. The graph connectivity is defined by the mesh topology of the garment, augmented with bi-directional edges.

PhysDrape is trained on a single NVIDIA V100 GPU for 150,000 iterations using the Adam optimizer \cite{zhang2018improved} with a initial learning rate of $5 \times 10^{-5}$. 
Consistent with the self-supervised protocol of DrapeNet \cite{de2023drapenet}, training is conducted exclusively on static garment meshes in the canonical T-pose. 
Although our method supports variable material properties (e.g., stiffness and density), these parameters are aligned with the fixed settings from DrapeNet during quantitative comparison to ensure a fair evaluation. 
End-to-end training is performed with the learnable solver set to $T=3$ iterations, while for evaluation, we increase this to $T=15$ to demonstrate the consistent performance improvements achieved with additional solver steps.

\subsubsection{Evaluation Metrics.} 
To quantitatively evaluate garment draping, we use common \cite{de2023drapenet} physics-based energy metrics, including strain ($E_\text{strain}$) for surface stretching and bending ($E_\text{bend}$) for folding smoothness. 
Lower energy values indicate a more relaxed and physically plausible state. 
Additionally, we assess collision handling using the Body-to-Garment (B2G) ratio \cite{de2023drapenet,li2022dig}, which measures interpenetration by calculating the ratio of garment surface area penetrating the body to total garment area. 
These metrics directly measure the physical plausibility and geometric validity of the draped garments.
% Evaluating unsupervised garment draping is challenging due to the stochastic nature of cloth dynamics; multiple draped states can be physically valid for a single pose. Therefore, relying solely on Euclidean distance is often insufficient \cite{de2023drapenet, santesteban2022snug}.
% Instead, we focus on physical energy and collision handling for quantitative evaluation, as they directly measure the physical plausibility and geometric validity of the draped garments. Following DrapeNet~\cite{de2023drapenet}, we report three physics-based energy metrics: strain ($E_\text{strain}$) quantifying surface stretching, bending ($E_\text{bend}$) reflecting folding smoothness. Lower energy values generally indicate a more relaxed and physically plausible state. Additionally, we report the Body-to-Garment (B2G) ratio to assess collision handling, calculated as the ratio of the surface area of garment faces penetrating the body to the total garment area.

\subsection{Comparisons with State-of-the-Arts}

\begin{table}[t]
    \centering
    \caption{PhysDrape quantitatively outperforms previous state-of-the-art methods, as demonstrated by lower physical energy metrics ($E_{\text{strain}}$, $E_{\text{bend}}$) and interpenetration ratios (B2G) on CLOTH3D testing set.}
    \label{tab:sota_comparison}
    \setlength{\tabcolsep}{8pt}
    \begin{tabular}{l c c c}
        \toprule
        Method & $E_{\text{strain}} \downarrow$ & $E_{\text{bend}} \downarrow$ & B2G \% $\downarrow$ \\
        \midrule
        DeePSD~\cite{bertiche2021deepsd} & 7.22 & 0.01 & 7.2\phantom{00} \\
        DIG~\cite{li2022dig} & 6.32 & 0.01 & 1.8\phantom{00} \\
        DrapeNet~\cite{de2023drapenet} & 0.43 & 0.01 & 0.9\phantom{00} \\
        \midrule
        \textbf{PhysDrape ($T=3$)} & 0.20 & \textbf{0.004} & \textbf{0.05} \\
        \textbf{PhysDrape ($T=15$)} & \textbf{0.15} & \textbf{0.004} & \textbf{0.05} \\
        \bottomrule
    \end{tabular}
\end{table}

\subsubsection{Quantitative Comparison.} PhysDrape outperforms three state-of-the-art approaches: DeePSD \cite{bertiche2021deepsd}, DIG \cite{li2022dig}, and DrapeNet \cite{de2023drapenet} on the same testing set from CLOTH3D, as reported in Table \ref{tab:sota_comparison}. As shown in the results, baseline methods DeePSD and DIG exhibit high strain energies ($>6.0$) and noticeable B2G ratios (7.2\% and 1.8\%, respectively), while DrapeNet retains a B2G ratio of 0.9\% mainly through its self-supervised collision avoidance design. The proposed PhysDrape ($T=15$) significantly outperforms these baselines, demonstrated by lower fabric energies (0.15 strain energy, 0.004 bending energy) and significantly improved interpenetration at a negligible level (B2G $\approx$ 0.05\%). The superiority of PhysDrape is mainly attributed to the integration of an explicit physical solver within the deep learning model, ensuring physical plausibility.

We note that other garment draping methods cannot be fairly compared due to being either template-specific or dynamic. Template-specific methods, such as SNUG \cite{santesteban2022snug} and GAPS \cite{chen2024gaps}, require learning specific garment templates, whereas PhysDrape does not rely on template learning and can generalize to unseen templates, making a fair comparison infeasible. 
Dynamic methods, such as Hood \cite{grigorev2023hood} and SENC \cite{liao2024senc}, simulate fabric deformation as the body moves from a T-pose to a given pose, with motion patterns affecting the draping outcome. In contrast, PhysDrape focuses on a static body pose, without considering body motion, and its optimization objective differs from that of dynamic methods, preventing a fair comparison.

\subsubsection{Qualitative Comparison.} 
\begin{figure}[t]
    \centering
    \includegraphics[width=0.95\linewidth]{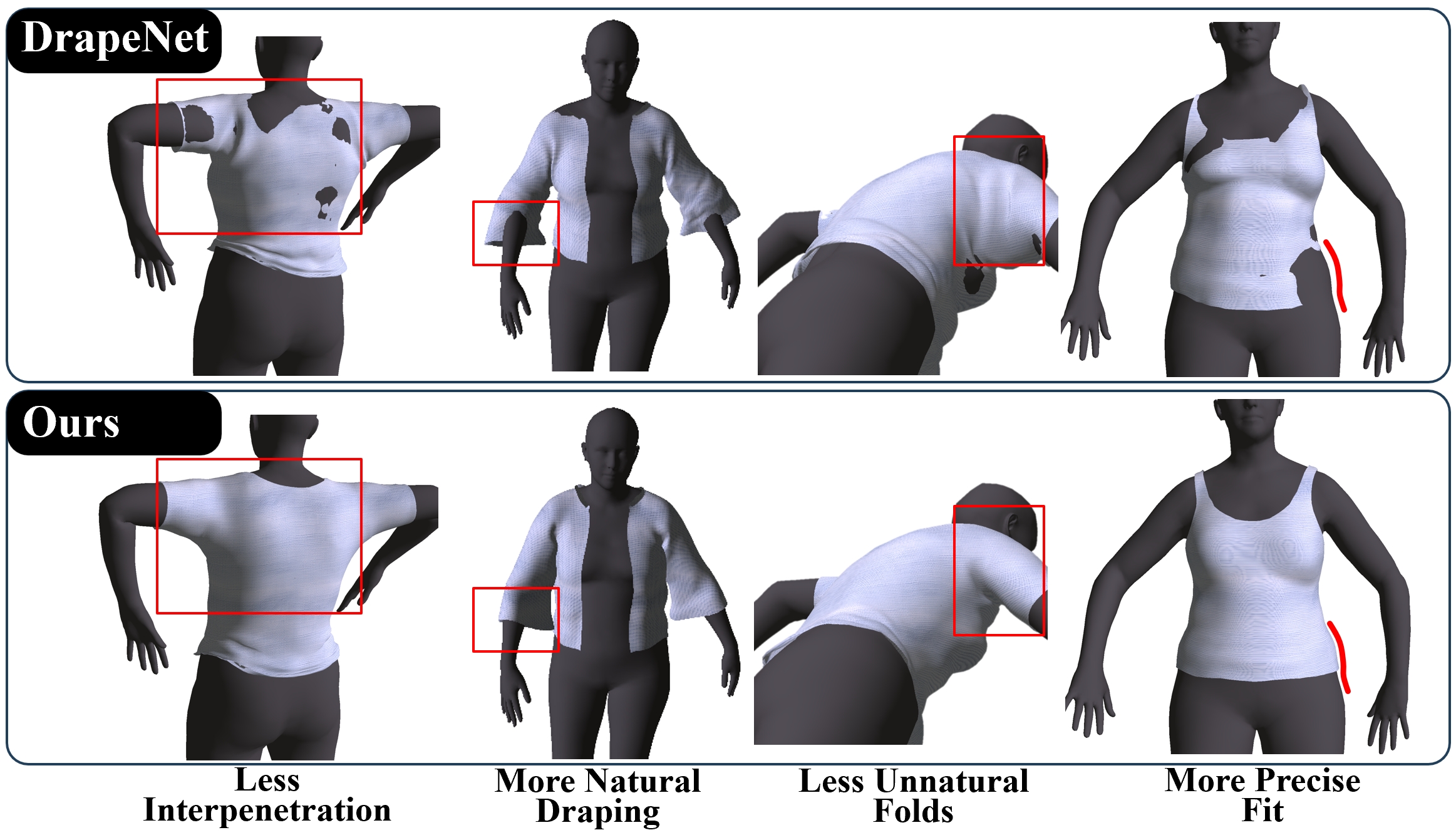}
    \caption{Garment draping visualization results demonstrating the superiority of PhysDrape over previous state-of-the-art, reflected in less interpenetration, more natural draping, fewer unnatural folds, and a more precise fit.}
    \label{fig:collision}
\end{figure}

Fig. \ref{fig:collision} demonstrates the visualization superiority of PhysDrape over DrapeNet. We present results across different body poses and garment types. As highlighted in red color, PhysDrape exhibits less interpenetration, more natural draping, fewer unnatural folds, and a more precise fit. Specifically, at areas with body curvature, PhysDrape produces garments that conform more naturally to the body’s contours, appearing more physically realistic. 
Additionally, at boundary regions, such as cuffs, collars, and underarms, most existing methods suffer from interpenetration, where the hands, neck, or shoulders penetrate the garment model. 
This issue arises primarily due to the lack of explicit physical constraints, as relying solely on neural network optimization makes it difficult to achieve accurate garment draping results.
In contrast, PhysDrape explicitly integrates physical solvers with deep learning models, ensuring more physically plausible garment draping results.

% We visually compare the draping quality against DrapeNet in Fig. \ref{fig:collision}. As highlighted by the red boxes, the baseline method suffers from severe artifacts, exhibiting explicit interpenetrations in complex contact regions such as the armpits, shoulders, and back. These failures typically occur in high-curvature or high-stress areas where soft penalties alone are insufficient to enforce physical constraints. In contrast, PhysDrape generates smooth, continuous, and collision-free garments that tightly conform to the body shape, validating the effectiveness of our hard-constrained projection in handling complex body-garment interactions.

\subsection{Ablation Studies}

To further demonstrate the effectiveness of the designs in PhysDrape, we conduct a series of ablation experiments, including: 1) the effectiveness of the three modules, 2) the convergence of multi-round simulations in the Stretching Solver, 3) acceptable runtime, 4) the effectiveness of joint optimization with the collision handler, and 5) the controllability of manually specified material stiffness. More ablation experiments are provided in the supplementary material.

\begin{table}[t]
    \centering
    \caption{Effectiveness of the three contributed modules: Force-Driven GNN, Stretching Solver, and Collision Handler in PhysDrape, measured by strain energy ($E_{\text{strain}}$), bending energy ($E_{\text{bend}}$), and interpenetration ratios (B2G) on the CLOTH3D dataset.}
    \label{tab:ablation_components}
    \setlength{\tabcolsep}{3pt}
    \begin{tabular}{c c c c c c}
        \toprule
       \makecell[c]{Force-Driven \\ GNN} & \makecell[c]{Stretching \\ Solver} & \makecell[c]{Collision \\ Handler} & $E_{\text{strain}} \downarrow$ & $E_{\text{bend}} \downarrow$ & B2G \% $\downarrow$ \\
        \midrule
        \checkmark & & & 0.20 & 0.004 & 3.50 \\ % 0.0032
        \checkmark & \checkmark & & 0.17 & \textbf{0.003} & 2.29 \\ % 0.0029
        \checkmark & & \checkmark & 0.21 & 0.004 & 0.06 \\ % 0.0041
        \midrule
        \checkmark & \checkmark & \checkmark & \textbf{0.15} & 0.004 & \textbf{0.05} \\ % 0.0036
        \bottomrule
    \end{tabular}
\end{table}

\subsubsection{Effectiveness of Modules.} Table \ref{tab:ablation_components} shows the contribution of the three main modules to the results. As seen in the table, the three modules effectively complement each other, leading to improved performance.
Specifically, directly using GNN for regression yields the worst results, as it lacks explicit physical guidance, and the regression results are not physically grounded. 
The stretching solver is responsible for force-based garment deformation. When used after the force-driven GNN, it reduces the internal energy of the fabric through stretching, making the result more plausible. However, without a collision avoidance mechanism, it still leads to a significant increase in the interpenetration ratio, especially within the self-supervised framework. 
The collision handler significantly reduces the interpenetration ratio by integrating physical constraints into the differentiable model in PhysDrape, achieving approximately a 95\% reduction in interpenetration.

%\begin{figure}[h]
%    \centering
%    \includegraphics[width=0.85\linewidth]{Figure5_2.png}
%    \caption{\textbf{Visualization of physical relaxation and force dissipation.} We map residual forces to a heatmap where red denotes high residual force. While DrapeNet exhibits widespread unnatural stretching, our solver effectively dissipates these forces from $T=3$ to $T=15$, reaching a stable and physically consistent equilibrium.}
%    \label{fig:heatmap}
%\end{figure}

\begin{figure}[t]
    \centering
    \includegraphics[width=0.8\linewidth]{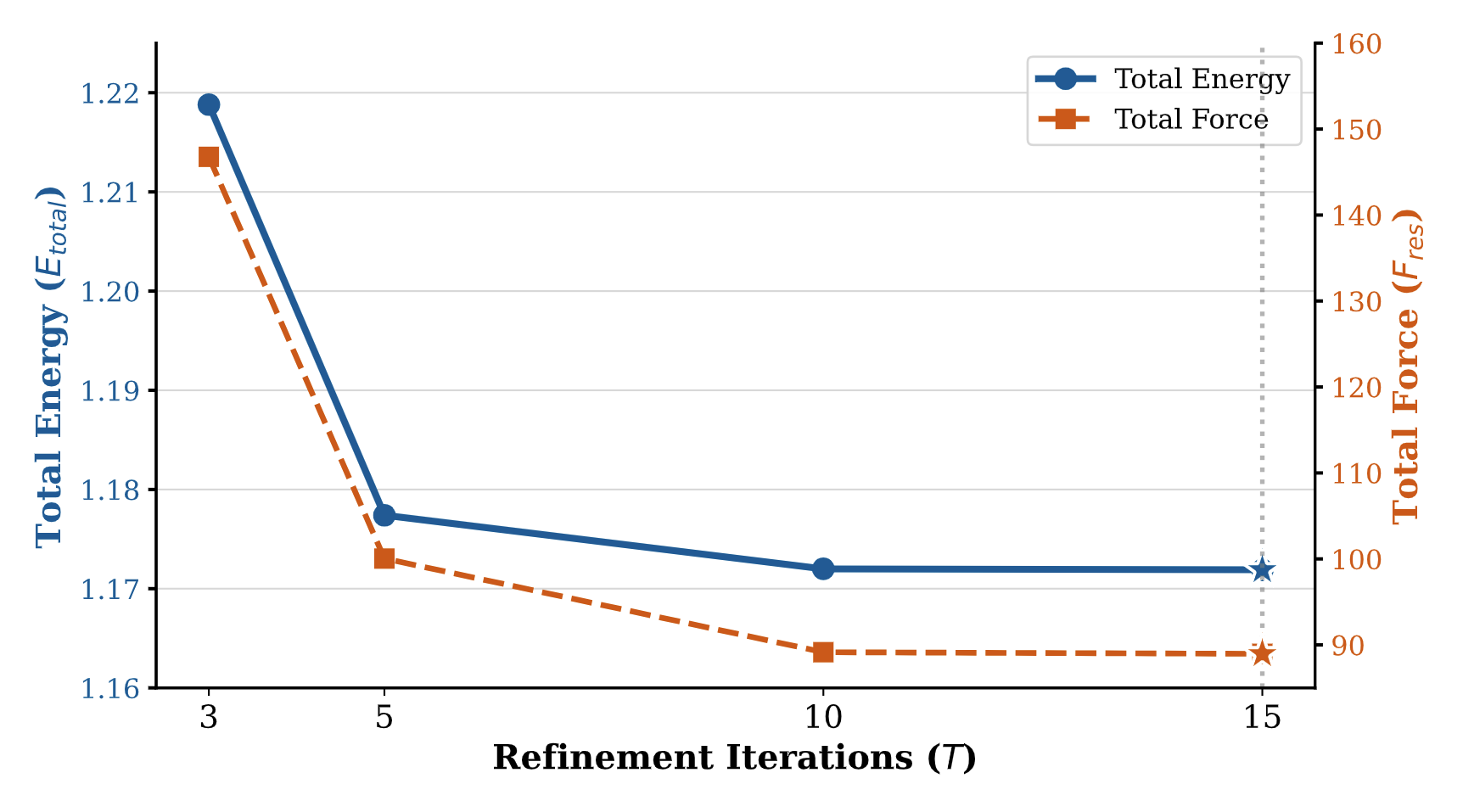}
    \caption{Total energy (blue line) and total force (orange line) converge as the stretching solver iterates, indicating the effectiveness of the physical solver. This demonstrates that the excess energy in the garment decreases as deformation progresses, and the system converges to a stable value without resulting in collapse.}
    \label{fig:curve}
\end{figure}

\subsubsection{Impact of Stretching Solver Iterations.} Fig. \ref{fig:curve} shows that total energy and residual forces converge to a stable value as the stretching solver iterates. This indicates that, on one hand, the model effectively reduces unnatural stretching, and on the other hand, the reasonable model design prevents structural collapse after multiple force-based stretching iterations.

% \subsubsection{Impact of Stretching Solver Iterations.} Fig. \ref{fig:curve} To validate the role of our two-stage coupling, we evaluate the performance at early ($T=3$) and converged ($T=15$) solver steps in Table \ref{tab:sota_comparison}. As shown in the B2G metric, our method achieves a near-zero B2G ratio of 0.05\% even at $T=3$, confirming that the projection layer efficiently handles the penetration. Extending iterations to $T=15$ further reduces the strain energy ($E_\text{strain}$ drops from 0.20 to 0.15). We visualize this physical relaxation process in Fig. \ref{fig:heatmap}. While DrapeNet exhibits widespread unnatural stretching (indicated by high residual forces shown in red), we observe that our method with $T=3$ produces significantly reduced residual forces that are localized. As the solver iterates to $T=15$, these remaining forces are effectively dissipated (turning blue), indicating that the mesh has relaxed to a stable state. This convergence behavior is quantitatively analyzed in Fig. \ref{fig:curve}, where both the total energy and residual forces drop significantly and stabilize, demonstrating that our solver reliably guides the mesh toward a lower-energy state.

\subsubsection{Runtime Assessment.} The runtime for draping each garment is ~29 ms when $T=3$ and ~91 ms when $T=15$ on a single NVIDIA V100 GPU. These results are obtained experimentally by averaging over 5000 garment draping samples.
According to classical user perception theory, 100 ms is the threshold for human perception \cite{time1968, time1994}, with responses below this time considered instantaneous. 
Therefore, whether at $T=3$ or $T=15$, PhysDrape achieves state-of-the-art results, as shown in Table \ref{tab:sota_comparison}, and can be regarded as a real-time system.

\begin{figure}[t]
    \centering
    \includegraphics[width=1.0\linewidth]{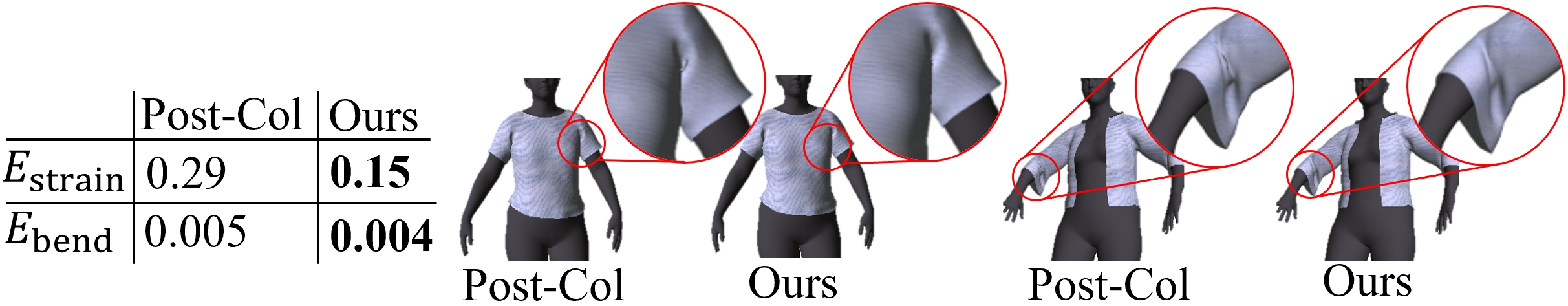}
    \caption{Compared to using independent collision avoidance postprocessing (Post-Col), our PhysDrape unifies the collision handler within a learnable model, resulting in improved detail accuracy and energy.}
    \label{fig:postcollision}
\end{figure}

\subsubsection{Impact of Optimizable Collision Handler.} Fig. \ref{fig:postcollision} illustrates that integrating collision avoidance into the optimization process of PhysDrape improves both the physical realism of garment details and reduces overall energy loss, compared to relying on independent postprocessing methods.

\subsubsection{Controllability of Material Stiffness.} 
\begin{figure}[t]
    \centering
    \includegraphics[width=1.0\linewidth]{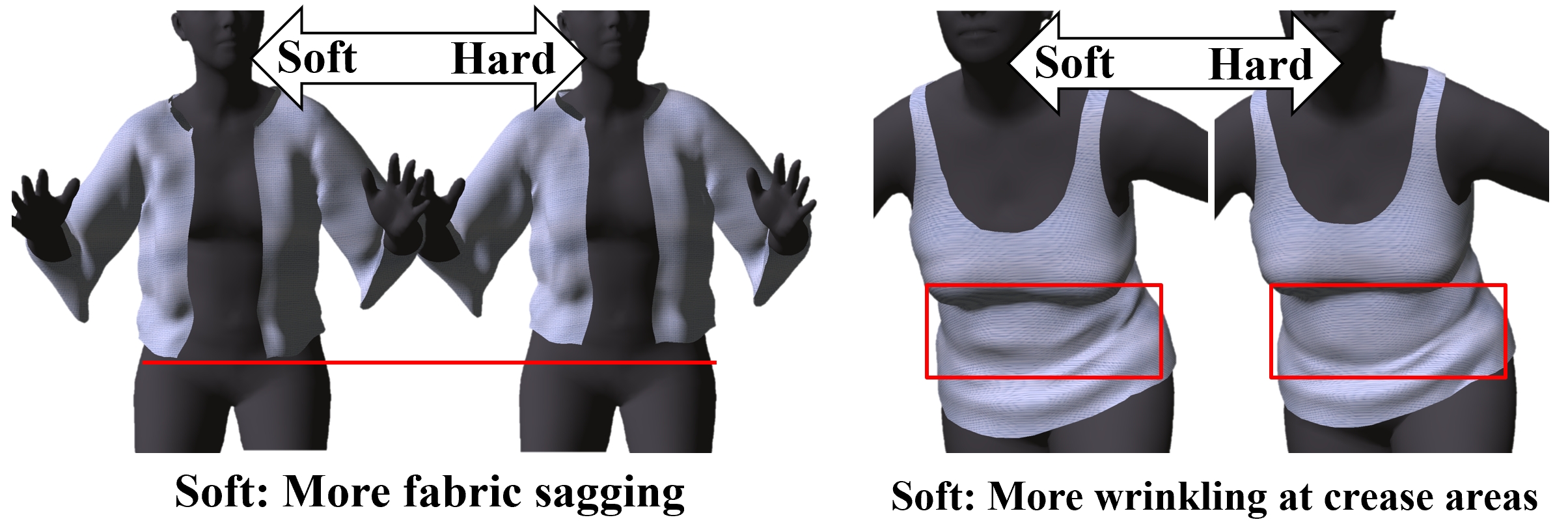}
    \caption{Visualization results show that material stiffness can be adjusted by manually changing material parameters. Softer materials exhibit more fabric sagging and more wrinkling, particularly at crease areas, consistent with observed physical behavior.}
    \label{fig:material}
\end{figure}

Fig. \ref{fig:material} demonstrates that PhysDrape allows the adjustment of fabric stiffness in garment draping by changing physical parameters. This level of control is difficult to achieve in other deep learning-based garment draping methods, highlighting the controllability and physical realism of the PhysDrape design. 
Specifically, stiffness can be controlled by adjusting $k_{\text{bend}}$ in Eq. \ref{eq:forces_bend} and the Lamé constants $\mu$ and $\lambda$ of $\mathbf{P}_T$ in Eq. \ref{eq:forces_strain}, with higher values resulting in stiffer fabrics. 
The results show that softer fabrics tend to sag more and exhibit more wrinkling at the crease areas, which aligns with physical observations. 
This illustrates that PhysDrape not only achieves better quantitative metrics through deep learning generalization but also incorporates physical solvers to generate more physically realistic details with strong controllability.

\section{Conclusion}
In this work, we presented PhysDrape, a novel garment draping framework that bridges physical modeling with deep learning to achieve physically realistic 3D garment deformation. 
The method consists of three learnable core modules: the force-driven GNN for predicting forces on each node, the stretching solver for modeling fabric deformation, and the collision handler for penalizing interpenetration. 
The entire framework supports self-supervised, end-to-end optimization and generalizes directly to unseen garment templates.
Extensive experiments on the CLOTH3D dataset demonstrate that PhysDrape outperforms previous state-of-the-art methods both quantitatively and qualitatively, while maintaining real-time inference. 
Ablation studies validate the effectiveness of each design, and visualization results further confirm the physical realism of garment draping as well as the controllable simulation of material stiffness.

%In this paper, we presented PhysDrape, a hybrid neural-physical framework that archieve physically realistic garment draping. Unlike existing methods that rely on soft penalties, which force a trade-off between geometric feasibility and physical plausibility, PhysDrape integrates a Physics-Informed GNN with a differentiable explicit solver. By coupling a learnable Force Solver based on the StVK model with a Differentiable Projection Layer, our pipeline reduces the interpenetration ratio to a negligible level (B2G $\approx$ 0.05\%) while minimizing residual forces to ensure physical fidelity. Extensive experiments demonstrate that PhysDrape achieves state-of-the-art performance, generating negligible interpenetration garments with significantly lower strain energy compared to baselines, all while maintaining real-time inference speeds. 

% \subsubsection{Limitations and Future Work.} PhysDrape resolves collisions in real-time but currently neglects inertial effects due to its quasi-static formulation. Future work will extend the solver to full animation and leverage differentiability for inverse garment reconstruction \cite{de2023drapenet}. Additionally, integrating implicit representations or sewing pattern optimization \cite{li2025dress, liu2023towards} shows promise for handling complex multi-layered garments.

%\section*{Acknowledgements}
%Please insert your acknowledgments here.

\bibliographystyle{splncs04}
\bibliography{main}
\end{document}